\documentclass[letterpaper, 10 pt, conference]{ieeeconf}

\IEEEoverridecommandlockouts
\usepackage{cite}
\usepackage{amsfonts}
\usepackage{amssymb}
\usepackage{graphicx}
\graphicspath{/Figures/}
\usepackage{amsmath}
\usepackage{epstopdf}
\usepackage{subfigure}
\usepackage{cite}
\usepackage{array}
\usepackage{url}
\usepackage{booktabs}
\usepackage{color}
\usepackage[percent]{overpic}
\usepackage[table]{xcolor}
\usepackage{tabularx}
\usepackage{multirow}
\usepackage{hhline}
\usepackage{fontenc}
\usepackage{dictsym}
\usepackage{mathabx}
\usepackage{xcolor}
\definecolor{Gray}{gray}{0.8}
\renewcommand{\baselinestretch}{0.95}
\let\labelindent\relax
\usepackage[inline]{enumitem}

\title{\LARGE \bf
Whole-Body Postural Adjustment Through Multi-Modal\\ Haptic Feedback Devices
}

\title{\LARGE \bf
Performance Analysis of Vibrotactile and Slide-and-Squeeze Haptic Feedback Devices for Limbs Postural Adjustment}

\author{Marta Lorenzini$^{1}$, Simone Ciotti$^{2}$, Juan M. Gandarias$^{1}$, Simone Fani$^{2}$,\\Matteo Bianchi$^{2,\dagger}$ and Arash Ajoudani$^{1,\dagger}$
\thanks{This work was supported in part by the European Union's Horizon 2020 research and innovation programme under grant agreement No.  871237 (Sophia), and in part by the Italian Ministry of Education and Research (MIUR) in the framework of
the CrossLab project (Departments of Excellence), and in the framework of PRIN (Programmi di Ricerca Scientifica di Rilevante Interesse Nazionale) 2017 with the project TIGHT: Tactile InteGration for Humans and arTificial systems (Grant number
818 2017SB48FP).}
\thanks{$^{1}$M. Lorenzini, J.M. Gandarias, and A. Ajoudani are with the HRI$^2$ Lab, Istituto Italiano di Tecnologia, 16163 Genoa (GE), Italy.
        {\tt\small marta.lorenzini@iit.it}}%
\thanks{$^{2}$S. Ciotti, S. Fani, and M. Bianchi are with the Information Engineering Department and the Research Center ``E. Piaggio'', University of Pisa, 56122 Pisa (PI), Italy.
        {\tt\small simone.ciotti@ing.unipi.it}}%
\thanks{$^{\dagger}$This work was equally supervised by these authors.}%
}

\begin{document}

\maketitle
\thispagestyle{empty}
\pagestyle{empty}


\begin{abstract}
  Recurrent or sustained awkward body postures are among the most frequently cited risk factors to the development of work-related musculoskeletal disorders (MSDs). To prevent workers from adopting harmful configurations but also to guide them toward more ergonomic ones, wearable haptic devices may be the ideal solution. In this paper, a vibrotactile unit, called ErgoTac, and a slide-and-squeeze unit, called CUFF, were evaluated in a limbs postural correction setting. Their capability of providing single-joint (shoulder or knee) and multi-joint (shoulder and knee at once) guidance was compared in twelve healthy subjects, using quantitative task-related metrics and subjective quantitative evaluation. An integrated environment was also built to ease communication and data sharing between the involved sensor and feedback systems. Results show good acceptability and intuitiveness for both devices. ErgoTac appeared as the suitable feedback device for the shoulder, while the CUFF may be the effective solution for the knee. This comparative study, although preliminary, was propaedeutic to the potential integration of the two devices for effective whole-body postural corrections, with the aim to develop a feedback and assistive apparatus to increase workers' awareness about risky working conditions and therefore to prevent MSDs.
\end{abstract}


\section{Introduction}
\label{Sec::Intro}
Work-related musculoskeletal disorders (MSDs) remain the most common work-related health problem in the European Union nowadays. Besides the harmful effects on workers themselves, they lead to high costs to enterprises, and society~\cite{EUOSHA2019}.
Extended research on this topic has identified several physical working conditions that may increase the risk of
developing MSDs~\cite{james2018global}. Recurrent or sustained awkward body postures are among the most frequently cited. The main affected body districts are the back, neck, shoulders, and upper limbs, but also disorders in the lower limbs have been reported~\cite{da2010risk}. 
To prevent the associated harmful effects, the study of systems that may correct the actions of workers during everyday industrial tasks has recently gained considerable interest. These systems may act as postural correction devices to reduce the excessive and continuous load on the joints due to non-ergonomic body configurations of the operators~\cite{punnett2004work}.

This complex issue has been approached by different disciplines starting with the design of ergonomic workstations~\cite{bossomaier2010scientific,ben2002ergonomic}. Despite this progress, the development of assistive devices to warn the workers about inappropriate postures has seen less attention. Some examples of assistive devices with similar goals can be found in literature, often proposing cumbersome solutions. In~\cite{lim2015development}, for example, a lower limbs exoskeleton was proposed to improve the muscle strength of the wearer while transporting loads. Other solutions, focusing on the control of collaborative robots to reduce human collaborators' excessive load, have been proposed~\cite{peternel2014teaching, evrard2009teaching}.

For their structure and characteristics, the introduction of postural warning systems in industrial environments is not trivial: the use of visual displays is not always suitable since it can act as a distraction \cite{diefenbach2021improving}. Meanwhile, the use of acoustic cues can be inefficient due to the noises present in the industrial settings \cite{goomas2010ergonomics}.

\begin{figure}[t]
    \centering
    \includegraphics[width=0.85\columnwidth]{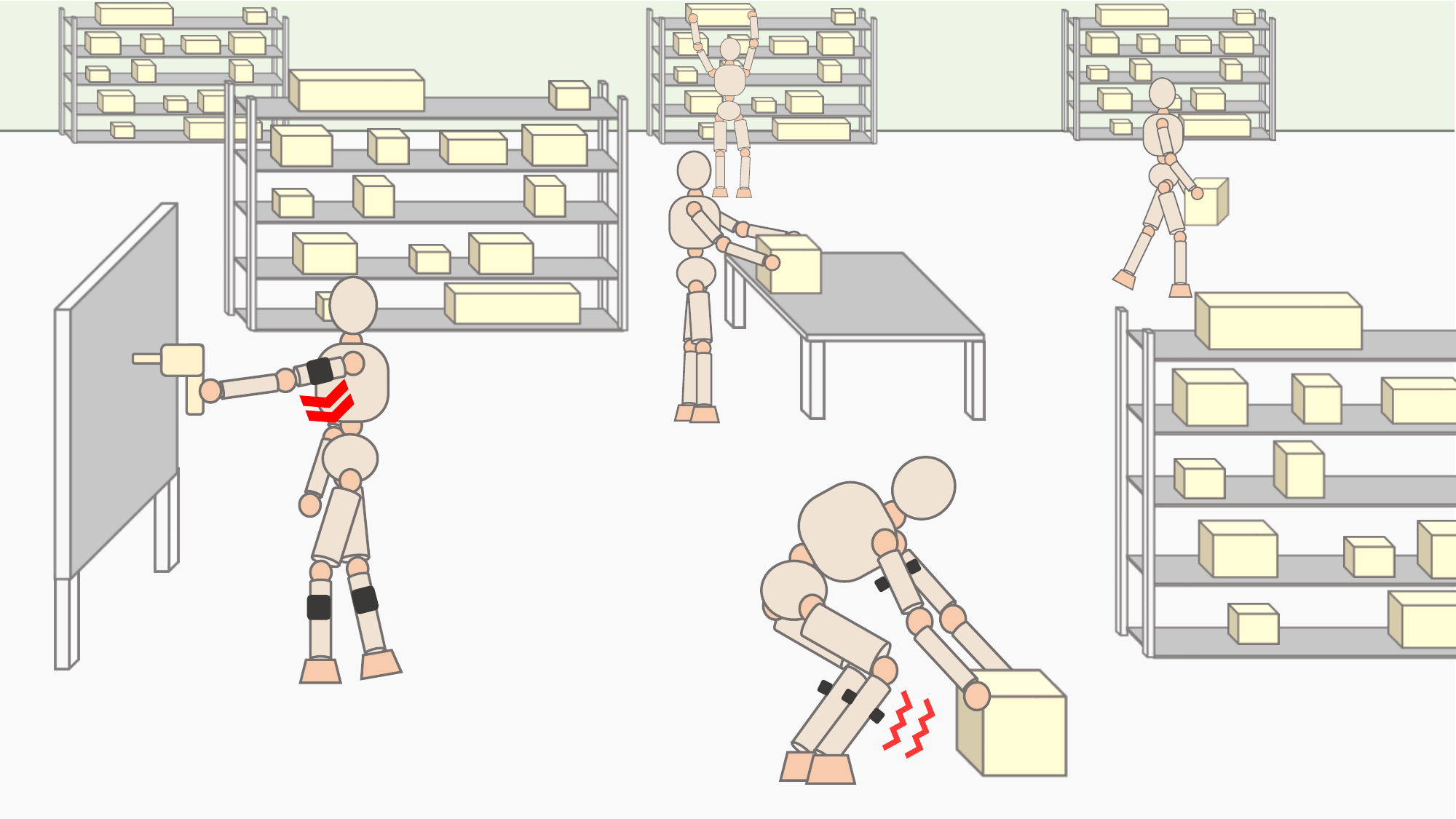}
    \caption{A view of a possible integration of the two haptic devices compared in the present manuscript.\label{Fig::Fin_Int}}
	\vspace{-0.5cm}
\end{figure}

On the other hand, in recent years, many wearable haptic devices have been proposed in a wide range of scenarios for human-human, and human-machine interaction \cite{pacchierotti2017wearable}. One of the strengths of these systems is the capability of being integrated with the operators' bodies and being easily carried around. The introduction of these devices allowed delivering different types of stimuli in a natural and unobtrusive way. Hence, their use to provide directional feedback has been widely investigated in different fields, ranging from guidance systems for blind people~\cite{barontini} to rehabilitation and training procedures~\cite{baldi2020hand,pezent2018separating}.

For their nature and characteristics, these devices can be used to provide directional feedback in applications of kinematic guidance for ergonomic enhancement. Some examples for this application can already be found in literature, as~\cite{rotella2012hapi}, in which a set of eccentric mass motors are used to provide vibrotactile feedback guidance of a static pose. 

Within this field, in two recent works, two systems for the postural correction of industrial workers have been presented by the authors of this paper. In Kim et al. 2021~\cite{kim2021directional}, a vibrotactile device called ErgoTac was used to develop three different feedback modalities to provide directional guidance at the level of the single joints towards the desired pose. On the other hand, in Fani et al. 2021~\cite{fani2021multi}, a wearable multi-cue system to be worn on the upper limbs was used to provide corrective feedback for posture balancing, conveying both squeezing stimuli and vibration. 

In this paper, we compare the two previously presented wearable haptic devices in their capabilities of providing guidance cues for the control of single joint angles, considering upper- and lower-limb postures. For what concerns the system in~\cite{fani2021multi}, only the slide and squeeze stimulation modality of the device was here considered. More specifically, the two devices are used to guide participants' shoulder and knee angles in a controlled experimental setting and evaluated through quantitative metrics. After the experiments, participants also undergo a subjective evaluation with a 7-points Likert-Scale Questionnaire and a NASA-TLX to evaluate their acceptance and physical and cognitive demand. 
An essential contribution of this work is implementing a modular environment where multiple sensors and feedback systems can be integrated, facilitating communication, data sharing, and synchronization. 

This study is headed toward the final  
optimised integration between the two systems, in which each device is placed on specific body segments maximizing the intuitiveness and the efficiency of the feedback system as a whole (see Figure~\ref{Fig::Fin_Int}).

The rest of the manuscript is organized as follows. 
In Section~\ref{Sec::Haptic} a description of the two used wearable devices with their guidance strategies and the integration setup is provided. The experiments are presented in Section~\ref{Sec::Experiments}, followed in Section~\ref{Sec::Results} by the presentation and discussion of the results. Section~\ref{Sec::Conclusions} concludes the manuscript, also exploring possible future applications and evaluations. 

\section{Multi-modal Haptic Feedback}
\label{Sec::Haptic}
The main goal of this work is to compare
two haptic devices previously presented in the literature, the ErgoTac \cite{kim2018ergotac} and the Clenching Upper-limb Force Feedback device (CUFF) \cite{fani2018simplifying}, in a postural correction setting. These two devices have been already previously used for postural and ergonomic correction  \cite{kim2021directional,fani2021multi}.
This section presents the mechanical structure and the main working principles of the two devices, followed by the description of the modular environment in which they are integrated. Next, the implementation of their haptic guidance is illustrated.

\begin{figure}
    \centering
    \includegraphics[width=0.80\columnwidth]{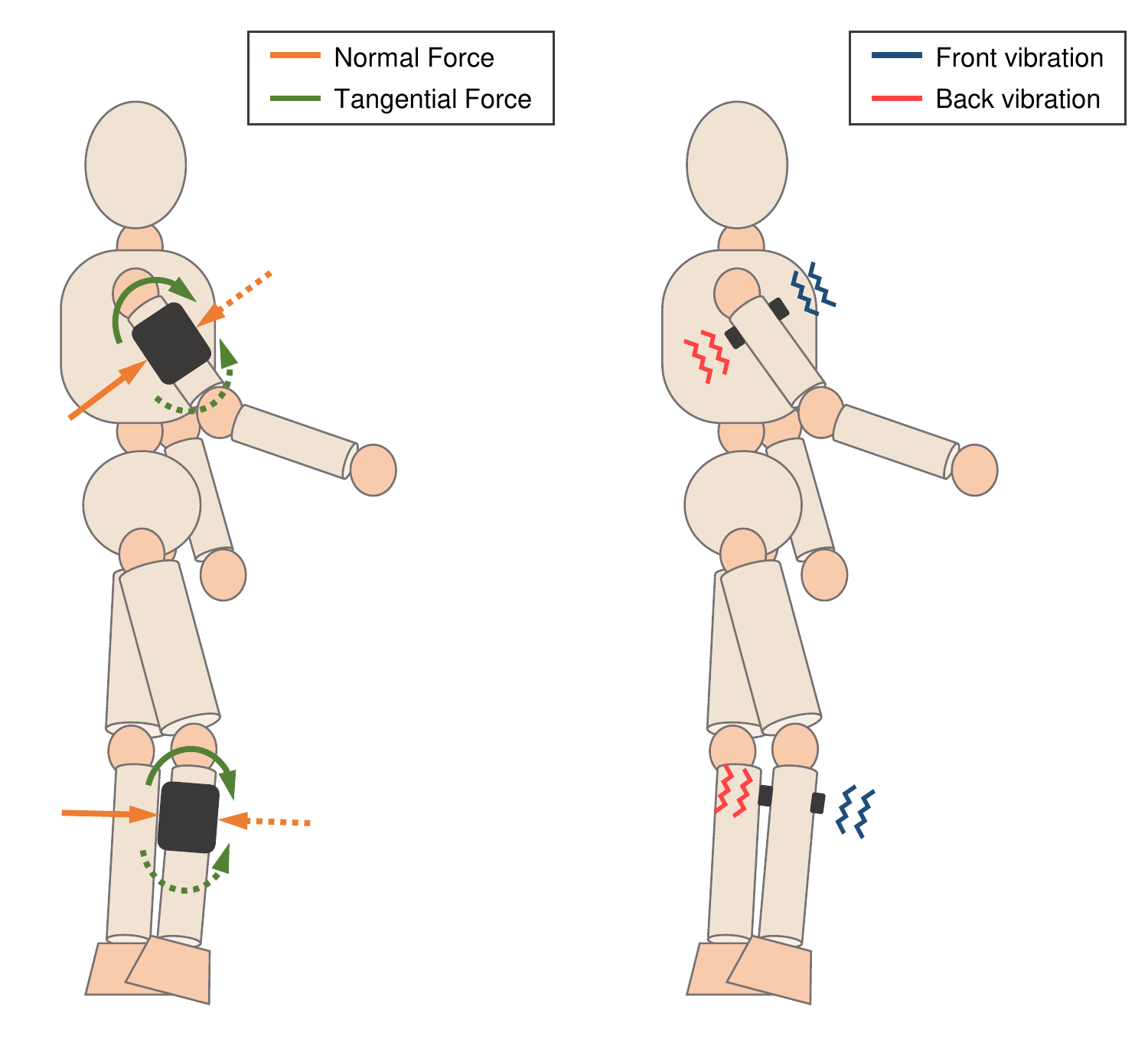}\\
    \hspace{-1.5cm} (a) \hspace{3cm} (b)
    \caption{Illustration of the feedback strategies adopted by the (a) CUFF and the (b) ErgoTac (SPOT modality).}
    \label{fig:modalities}
	\vspace{-0.5cm}
\end{figure}

\subsection{The CUFF}
The CUFF is a wearable haptic device initially designed as a force feedback tool for prosthetic application \cite{casini2015design} using a fabric-band actuated through two DC motors to provide normal and tangential force stimuli on the user's skin. For its structure, the CUFF can be easily worn on any limb segment through two velcro bands fixed to the main frame.

In the specific, the two DC motors fixed on the main frame of the device can be independently controlled to move the two extremities of the fabric band. If the two motors move in the same direction, a sliding movement of the fabric band is generated, applying a tangential force to the user's skin. On the other hand, if the two DC motors are controlled to move in the opposite direction, the fabric band, to be seen as a cuff wrapping the limb, can be squeezed or released, generating forces normal to the skin of the user (see Fig. \ref{fig:modalities}a). 
The fabric band, used as a user interface, is covered with a bio-compatible silicone layer, allowing it to generate higher tangential forces.

The control of the two DC motors is performed using a double control loop, one in current and one in position. This structure allows us to maintain precise and stable positions and better stabilize the grip of the fabric band on the users’ skin. The CUFF weights $\approx 230$~g, and its dimensions are $124\times70\times58$~mm.

The proposed device was already successfully applied in other fields such as telerobotics~\cite{fani2018simplifying}, rehabilitation robotics~\cite{pezent2018separating, pezent2019spatially} and assistive applications~\cite{barontini}. The CUFF is controlled using an onboard custom control board capable of managing the two motors' movements and communicating with the computer through bus RS485. 
In this work, differently from~\cite{fani2018simplifying} in which vibration was also included, we considered the CUFF in its original version. This choice was motivated by the need for clearly distinguishing the tactile cues delivered by the two devices considered in our work.

\subsection{The ErgoTac}
The ErgoTac is a wireless wearable vibrotactile feedback device that has been designed to be placed on the main joints of the human body to warn the users about possible overloading of the joints \cite{kim2018ergotac}. 

The single ErgoTac module weights only 28~g , and its dimensions are $68.1\times37.0\times17.3$~mm. Multiple modules can be combined in the different joints to obtain a distributed correction devices network on the user's body. To simplify the placement of the modules on the skin, the shape of the bottom surface presents a curved shape to better adapt to the human body segments. The different modules are kept in place through elastic bands (see Fig.~\ref{Fig::ExpSetup}b).
The use of the wireless communication (Bluetooth Low Energy in the 2.4~GHz spectrum), together with the integration of microcontroller and battery on the modules 
allows an increase in wearability, reducing also the obstacles to the user's movements.

The vibration is generated through a mini eccentric rotating mass (ERM) vibration motor, controlled in Pulse Width Modulation (PWM) with the microcontroller. The ErgoTac provides information to the user through three different levels of vibration amplitude, ``High'' (100\%), ``Medium'' (60\%), and ``Low'' (30\%), maintaining a constant frequency of the vibration at 121~Hz.

This device has been previously validated by a user study during which subjects reconfigured body posture to minimize overload efforts while performing a heavy lifting task \cite{kim2018ergotac}. Furthermore, in the same study, three vibration modalities (PATTERN, SPOT, and RAMP) have been considered to determine the best feedback modality from objective and subjective results provided by the subjects. The results yielded strong evidence on the usefulness and the intuitiveness of one of the developed modalities (the SPOT modality) in guiding towards ergonomic working conditions by minimizing the effect of an external load on body joints. In this study, the SPOT modality of the ErgoTac will be used. The SPOT modality, as represented in Fig.~\ref{fig:modalities}b, uses two ErgoTac units per joint: information on the desired direction is provided through repulsive vibration feedback on the opposite side of the limb (see~\cite{kim2021directional} for more detail).

\subsection{The Integration: ROS-based environment}

\begin{figure}
    \centering
    \includegraphics[width=0.9\columnwidth]{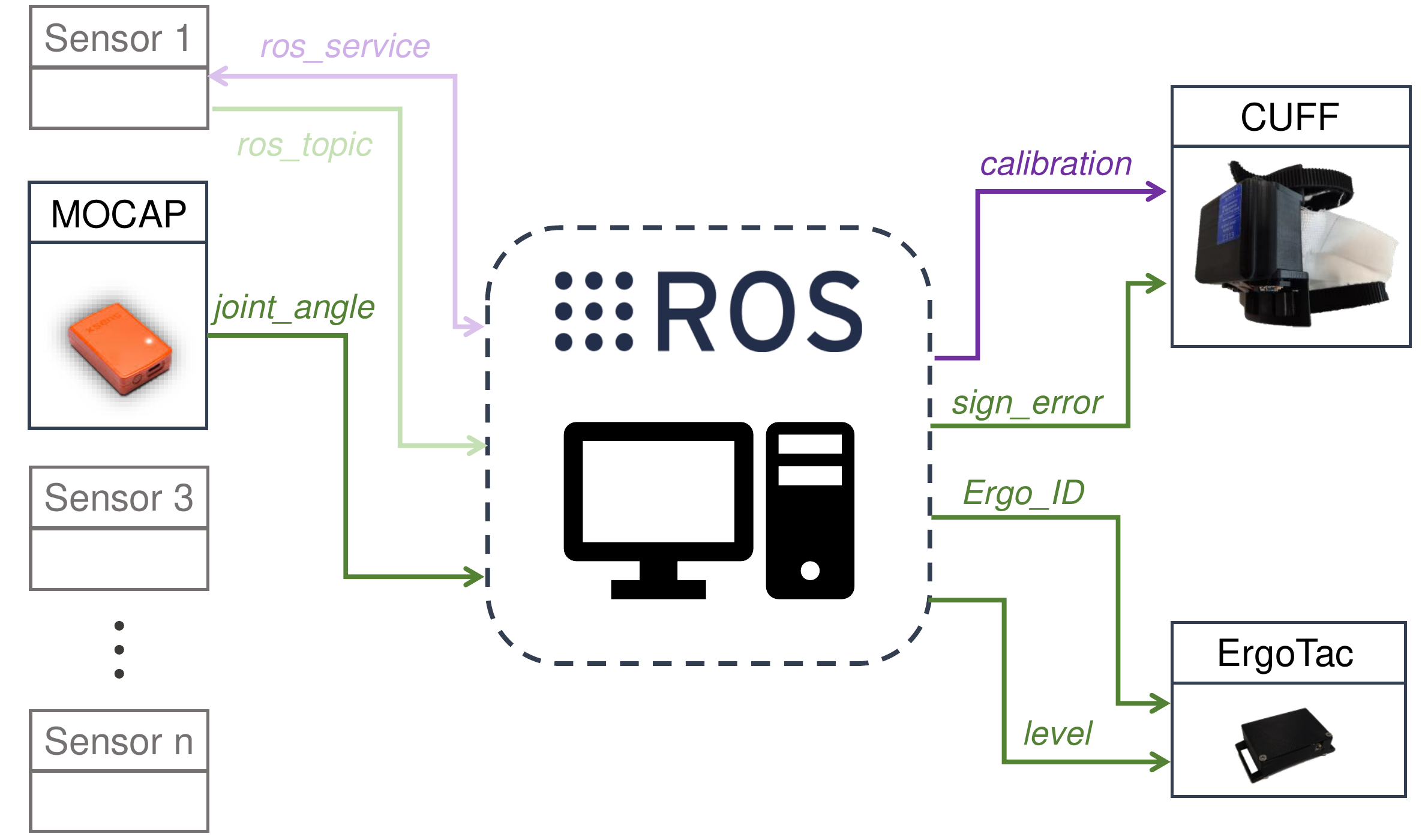}
    \caption{The multi-modal integration into the ROS middleware framework allows the data processing and communication between multiple sensor systems and haptic feedback devices.}
    \label{Fig::ros_interface}
	\vspace{-0.5cm}
\end{figure}

An essential part of this work is the integration of the involved devices within the same environment to allow an interchangeable setup in which they can be switched one with others allowing the system to work continuously, without the need for significant changes. To this aim, the tools offered by ROS (Robot Operating System\footnote{\url{https://www.ros.org/}}) are exploited. ROS-based communication interfaces are implemented for both the sensor and feedback systems employed in this work. Hence, a ROS-based environment is built. All the devices included can communicate and share data using ROS messages through ROS topics or ROS services, representing the many-to-many or the one-to-one communication paradigm.
Moreover, the data synchronization can be easily implemented through ROS time instances, enabling collective data storage. As illustrated in Fig. \ref{Fig::ros_interface}, the implemented environment is designed as a modular framework. Multiple devices (both sensor and feedback systems) can be added and share data with a central processor and/or between each other through ROS. In this work, an inertial-based motion-capture (MoCap) system is employed to collect data about human motion and share them with a central processor that, in turn, send commands to the two considered haptic devices. Specifically, information about the human joint angles are sent through ROS topics. A ROS service is implemented to calibrate the CUFF while ROS topics are used to control the haptic guidance of both the CUFF and ErgoTac. However, other sensor/feedback systems could replace and/or be added to the environment, with the only requirement of a ROS-based communication interface.     

\subsection{The Haptic Guidance Implementation}
The two haptic devices are used, in this work, to guide one or two human joints. The selected joints are the shoulder, commonly involved in manipulation tasks, and the knee, fundamental for whole-body posture. 

When the ErgoTac is used, two units are placed, one in the front and one in the back of the selected area. The SPOT modality is implemented. In this modality, the vibration is activated in the unit placed on the opposite part of the limb w.r.t. the movement to perform, acting as a repulsive cue. The three-level of vibration amplitude that ErgoTac can deliver (i.e., "High," "Medium," "Low") provide instead information on the amount of error made by the user during the movement.
The higher the vibration, the higher the error, i.e., the difference between the actual joint angle and the target one. Accordingly, the information required from ErgoTac is the unit ID and the vibration level (see Fig.~\ref{Fig::ros_interface}). The guidance provided by ErgoTac is sequential, meaning that only one unit is activated at a time. As such, when two joints should be guided, one joint is assisted at a time. 

When the CUFF is used, the device is worn on the biceps and/or on the lower part of the calf. 
To provide both intensity and direction of the correction, the information required from the CUFF is the signed error (see Fig.\ref{Fig::ros_interface}). 
The direction of the postural correction is provided with a tangential force in the direction of the correction. I.e., if the movement to perform is a forward movement, the CUFF will rotate clockwise, providing a forward tangential force on the internal part of the arm. On the contrary, if the movement to perform is backward, a counter-clockwise rotation of the cuff's fabric is generated, producing a backward tangential force on the skin of the internal part of the arm. The other degree of freedom of the CUFF, the squeezing, is instead used to provide information about the amplitude of the error. When the subject is placed in the goal position, and the error is null, the force exerted is $3$~N. If the user moves from the goal position, the CUFF starts squeezing with a force intensity proportional to the error with a maximum squeezing force of $20$~N when the error modulus is equal to or larger than $90^\circ$ (see \cite{fani2021multi} for more detail on the force limits). Combining the two stimuli generates a localized pinch on the side of the desired movement with an intensity proportional to the intensity of the error, i.e., in the front of the arm for forwarding correction movements and the back of the arm for backward correction movements. Unlike ErgoTac, two CUFF can be activated simultaneously.

\section{Experimental Analysis}
\label{Sec::Experiments}
The experiment's goal is to provide an idea about the performance and acceptability of the different haptic modalities and compare their intuitiveness. First, the experimental setup and protocol are described. Next, the evaluation tools are illustrated.

\subsection{Experimental setup}

\begin{figure}[t]
    \centering
    \includegraphics[trim=1cm 1cm 1cm 1cm,clip,width=0.8\columnwidth]{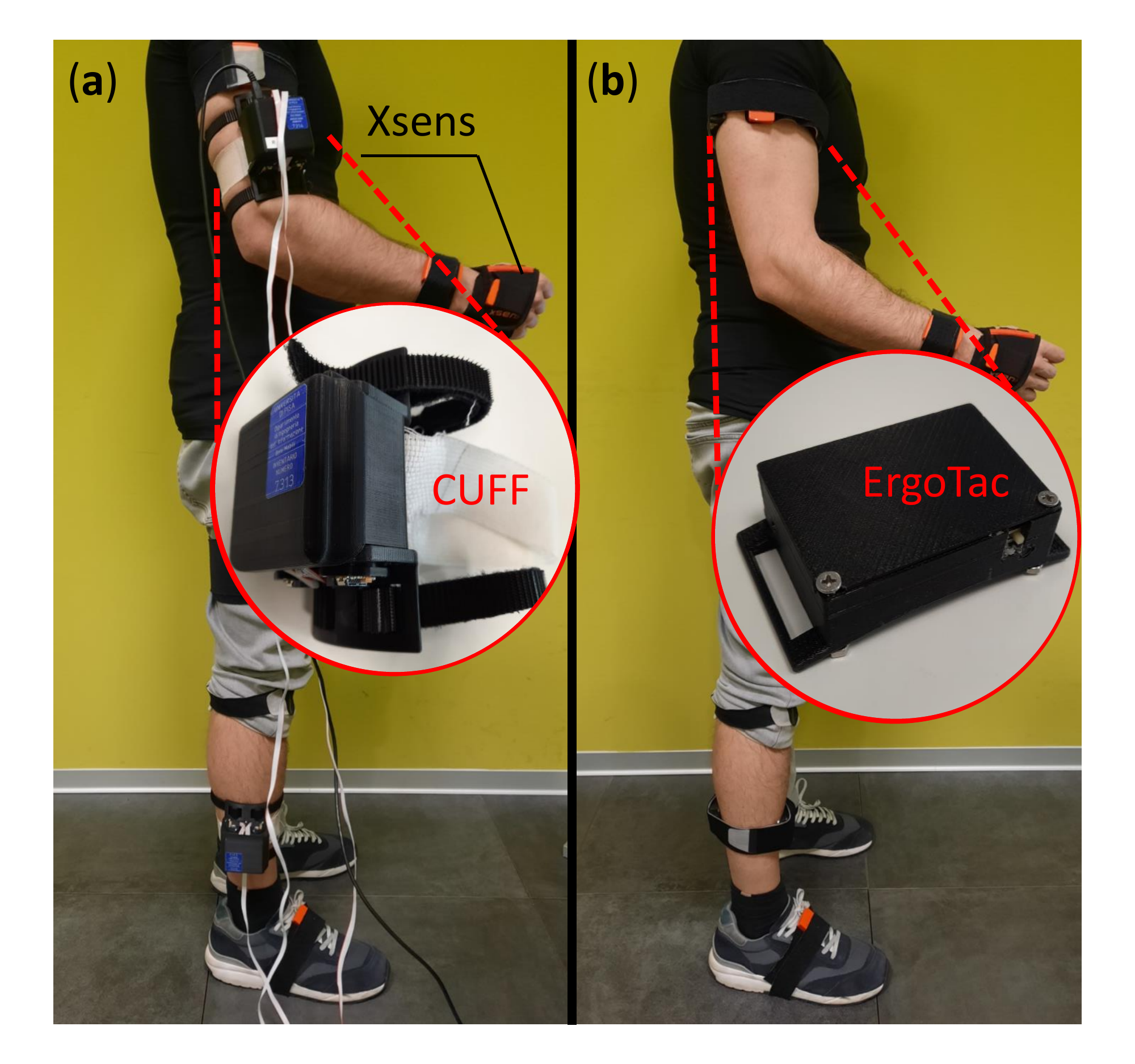}
    \caption{Overview of the experimental setup. The two haptic devices compared in this study are highlighted: the CUFF (a) and the ErgoTac (b).}
    \label{Fig::ExpSetup}
	\vspace{-0.3cm}
\end{figure}

The whole experimental procedure was carried out at Human-Robot Interfaces and Physical Interaction (HRII) Lab, Istituto Italiano di Tecnologia (IIT), and the protocol was approved by the ethics committee Azienda Sanitaria Locale (ASL) Genovese N.3 (Protocol IIT\_HRII\_SOPHIA 554/2020).
Twelve self-assessed right-handed volunteers (one female) took part in the experiment (age $28.3 \pm 2.4$ years). No participant presented physical or psychological impediments that could affect their capabilities of understanding and performing the tasks. The experimental setup is shown in Fig.~\ref{Fig::ExpSetup}. To collect information on human kinematics, an inertial-based MoCap system, the MVN Biomech suit (by Xsens Technologies BV, Enschede, Netherlands) was used.
The two haptic devices were used, one at a time, to guide one or two human joints, i.e. the shoulder, the knee or their combination. The right side of the body was chosen for systems placement hence the systems were placed on the right upper arm and/or on the right lower part of the calf, 
in standing position.
The haptic guidance was defined by directly comparing the subjects' current body configuration provided by the MoCap with a target one.

\subsection{Protocol}
Each participant was asked to follow the guidance provided by the haptic devices moving the appropriate body joint up to the reference angles.
Each experimental session was composed of two blocks, one using the CUFF and the other using the ErgoTac, respectively, as haptic guidance feedback. The order of the two blocks was randomized per participant. Each block was composed of three sub-blocks randomized within each block: i) upper limb guidance (focusing on the shoulder joint), ii) lower limb guidance (focusing on the knee joint), and iii) multi-joint guidance (focusing on both the shoulder and the knee joints at once).

Three reference angles on the sagittal plane were chosen for the shoulder and the knee, and three couples of angles were chosen for the combination of shoulder and knee; the selection was based on previous literature \cite{kim2021directional}. The chosen angles were $-10^\circ$, $45^\circ$, and $90^\circ$ for the shoulder and $30^\circ$, $80^\circ$, and $115^\circ$ for the knee. The couples $\left[20^\circ 110^\circ\right]$, $\left[55^\circ 70^\circ\right]$, and $\left[100^\circ40^\circ\right]$ were chosen for the multi-joint sub-block (first angle referring to the shoulder). 
Within each sub-block, the order of the single angle/couple of angles was also randomized.

As in the ``directional vibrotactile feedback modalities'' experiment in \cite{kim2021directional}, each trial concluded when the participant stated to be in the position goal. Additionally, in this work, we added a 90~s limit per trial; once the limit is reached, the trial was considered failed. The limit was chosen considering the average time and the standard deviation in the SPOT condition for the shoulder in \cite{kim2021directional}.
Before starting each block, a 5 minutes training session was performed. This session had the goal of allowing the participant to familiarize with the cues provided by the devices.

\subsection{Measurements and indices}
For the evaluation of the performances, the same indices previously used in \cite{kim2021directional} and reported in Table~\ref{Table::Ind} were used. The selected indices allowed an evaluation of both physical and cognitive aspects.

\begin{table}[t]
\begin{center}
\caption{Measurements and indices used for the systems evaluation. Refer to \cite{kim2021directional} for more detail.\label{Table::Ind}}
\rowcolors{2}{Gray}{white}
\begin{tabularx}{\columnwidth}{llc}
\toprule
Index*& Description &Unit\\
\midrule
Confusion index  & \begin{tabular}{l}
Percentage of time the subject goes in\\
the opposite direction of the guidance 
\end{tabular} &  \%    \\
Success ratio     & \begin{tabular}{l}
Percentage of cases where \\ the desired position was reached
\end{tabular}       & \%    \\
\midrule
Reaching time     & \begin{tabular}{l}
Duration of the reaching movement \\ to the desired position
\end{tabular}  & s     \\
Angular distance  & \begin{tabular}{l}
Total travelled joint angular distance \\ to the desired position
\end{tabular}      & deg   \\
Reaching~velocity & \begin{tabular}{l}
Average velocity from the initial \\ to the desired position
\end{tabular}        & deg/s \\
\midrule
Final error       & \begin{tabular}{l}
Percentage difference between the \\ final and desired position
\end{tabular}  & \% \\ \bottomrule
\end{tabularx}
\end{center}
*The first group is for all trials, the second for the cases with success, the third for the failed trials.
\vspace{-0.5cm}
\end{table}

After performing each experimental block, participants were asked to fill a 7-Points Likert-Scale questionnaire (1: strongly disagree; 7: strongly agree) and a NASA-TLX questionnaire \cite{hart1988development} on the specific device used in the block. A final 7-Points Likert-Scale questionnaire including comparative questions was then proposed at the end of the experiment. The NASA-TLX questionnaire allows getting a subjective self-evaluation of six parameters, i.e. 
Mental Demand (MD), Physical Demand (PD), Temporal Demand (TD), Performance (P), Effort (E), and Frustration (F).
All scores range from $0.0$ to $100.0$. Details about the Likert-Scales are instead provided in Table \ref{Tab::LT}. Statistical analysis was carried out on the performance indexes and the questionnaires' outcomes to check the significance of the results. The non-parametric Wilcoxon signed-rank test was conducted to compare CUFF against ErgoTac for each sub-block and the single joint against multi-joint condition for both the devices. 

\section{Results and Discussion}
\label{Sec::Results}
\begin{figure}[t]
    \centering
    \includegraphics[trim=1.5cm 0cm 3cm 0cm,clip,width=\columnwidth]{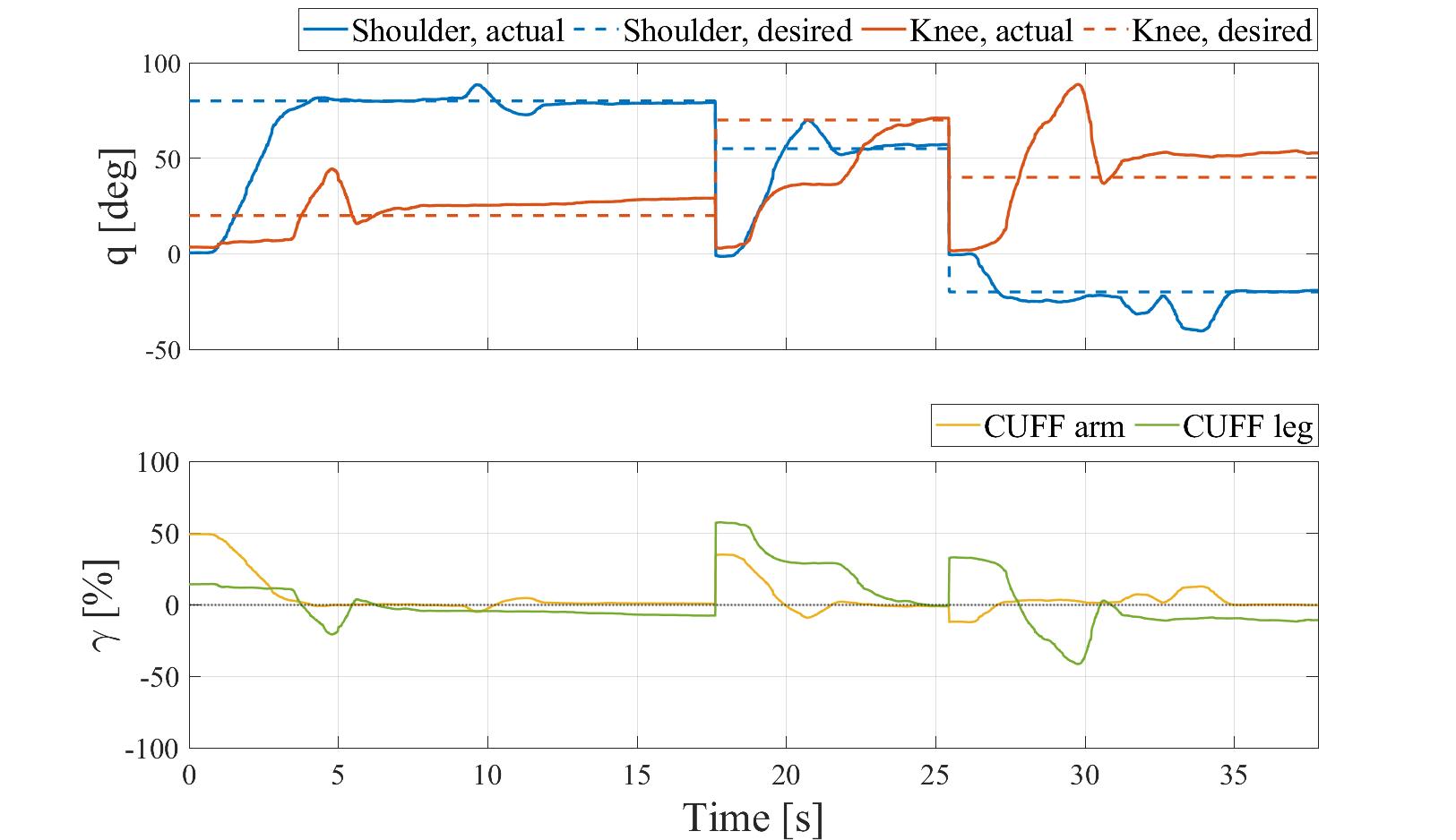}
    \caption{Results of the multi-joint sub-block (shoulder and knee joint guidance at once) with the CUFF feedback for one subjects. 
    }
    \label{Fig::Res_CUFF}
	\vspace{-0.1cm}
\end{figure}

\begin{figure}[t]
    \centering
    \includegraphics[trim=1.5cm 0cm 3cm 0cm,clip,width=\columnwidth]{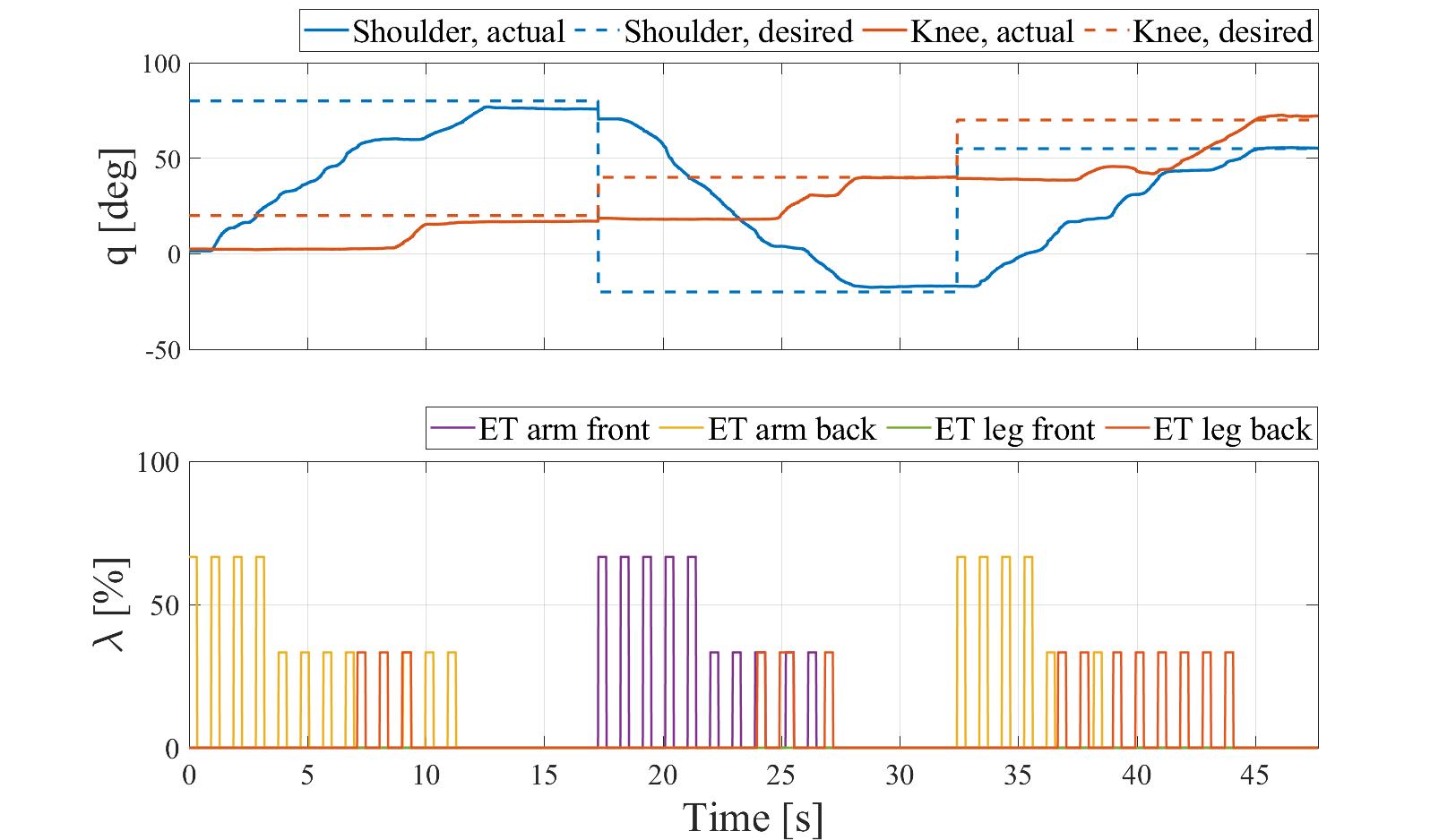}
    \caption{Results of the multi-joint sub-block (shoulder and knee joint guidance at once) with the ErgoTac feedback for one subjects. 
    }
    \label{Fig::Res_ET}
	\vspace{-0.5cm}
\end{figure}

Figures \ref{Fig::Res_CUFF} and \ref{Fig::Res_ET} represent the results of the multi-joint sub-block, in which the guidance was provided to the shoulder and the knee joint at once, with the CUFF and the ErgoTac, respectively, for one selected subject. The couples of angles $q$ were tested one at a time; however, the separate trials are here concatenated for synthesis. In these plots, the different functioning of the two devices can be clearly appreciated. The DC motors position $\gamma$ and the unit vibration level $\lambda$ are represented for the CUFF and the ErgoTac, respectively. While the CUFF provided a continuous stimulus, the ErgoTac generated subsequent pulses. As already mentioned, the two CUFFs (on the arm and the leg) can be activated simultaneously. In Fig. \ref{Fig::Res_CUFF}, it can be observed that the subject was able to follow the guidance of both the CUFFs at the same moment. On the other hand, the sequential feedback of ErgoTac guided the subject joints one at a time but still allowed them to reach the desired joint position.

\begin{figure*}[t]
    \centering
    \includegraphics[trim=1cm 1cm 1cm 0cm,clip,width=0.861\textwidth]{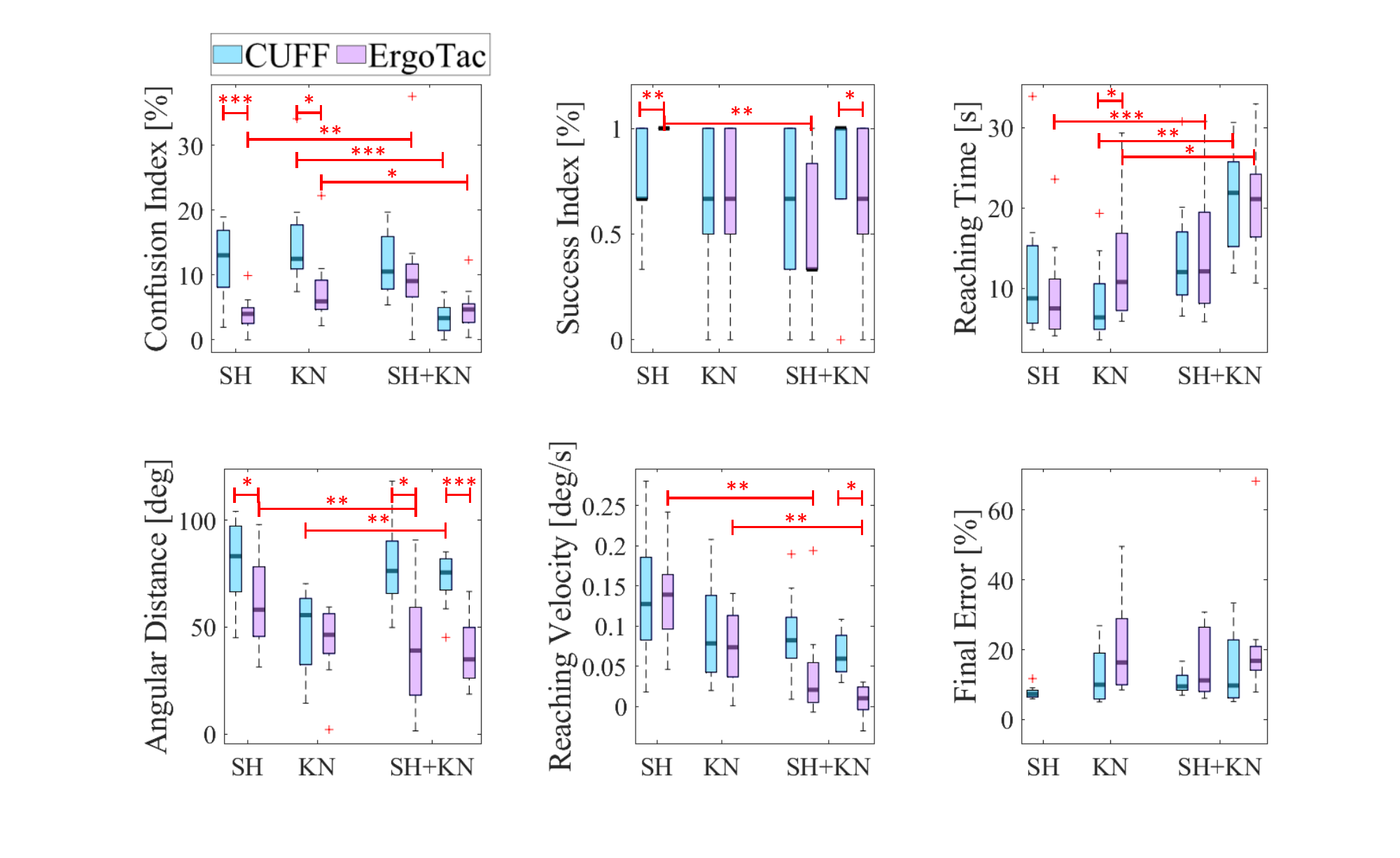}
    \caption{Results of the performance indices reported in Table \ref{Table::Ind} to compare CUFF and ErgoTac. The three experimental sub-block (SH: shoulder, KN: knee, SH+KN: shoulder and knee at once) are represented, averaged among all the trials and subjects. Significance level are indicated at the *p\textless{}0.05, **p\textless{}0.01, ***p\textless{}0.001.}
    \label{Fig::Boxplots}
\end{figure*}

In Fig.\ref{Fig::Boxplots}, the results of the performance indices used to compare the two devices are represented in boxplots. The indices scores are averaged among all the trials (i.e., angles/couples of angles tested within each sub-block) and among all the twelve subjects. The results of the statistical analysis are also illustrated, highlighting when the comparisons led to a p-value (p)~\textless{}0.05.

According to the confusion index results, when a single joint was guided, ErgoTac proved to be significantly less confusing in suggesting the direction of the postural correction for both the knee and the shoulder. 
Interestingly, with both the devices, the guidance on the knee was significantly less confusing in the multi-joint trial. 
Mixed results can be observed regarding the percentage of success. However, mainly in single-joint trials, both the devices proved to be effective in making the subjects complete the task. The reaching time significantly increased when two joints were guided (except for the shoulder with the CUFF). Considering instead the knee-only guidance, the CUFF allowed to significantly reduce the reaching time. On the other hand, the angular distance was significantly lowered when using the ErgoTac against the CUFF. As with the confusion index, a significant improvement can be observed in the angular distance for the shoulder in the multi-joint trial when using the ErgoTac. 
The reaching velocity significantly increased in the single joint guidance against the multi-joint one, presenting significant differences for the ErgoTac. On the other hand, the CUFF in the multi-joint guidance enabled a significantly improved reaching velocity for the knee. Lastly, no significant differences were found for the final error. 

Overall, the ErgoTac  demonstrated better performance in the single-joint guidance of the shoulder with improvements in 6/6 indices (3/6 significant). 
For what concerns the single joint guidance of the knee, there is evidence that the CUFF can induce faster responses and higher accuracy in reaching the desired position, and hence appearing as the effective solution.

When two joints were guided simultaneously, the performance of both the devices got worse (confusion index, reaching time and velocity, angular distance). Unexpectedly, the opposite was true in some cases, and the related aspects should be investigated. Nevertheless, a good percentage of success was reached, and the final errors in the case of failure were comparable to the single-joint case. This proves the potential of the devices in providing multi-joint guidance and encourages further studies on their use in whole-body applications. In particular, based on the single-joint results, their combination, i.e., ErgoTac for the shoulder and the CUFF for the knee, may produce good results.  

\begin{figure}[t]
    \centering
    \includegraphics[width=0.6\columnwidth]{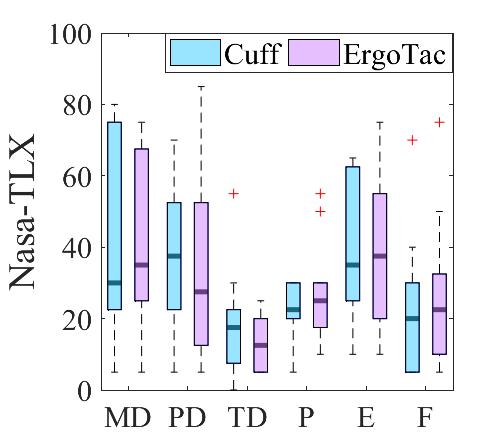}
    \caption{Results of NASA-TLX to compare CUFF and ErgoTac. The scores are averaged among all the repetitions and all the twelve subjects.}
	\label{Fig::NASA}
	\vspace{-0.5cm}
\end{figure}

In Table~\ref{Tab::LT}, the results of the Likert-Scales averaged among all the twelve subjects are provided in the right column (mean $\pm$ standard deviation). Fig.~\ref{Fig::NASA} represents instead the outcome of the NASA-TLX in boxplots. No significant differences were found in any of the questions' comparisons. 
However, a slight preference was given to the CUFF for the knee but to the ErgoTac for the shoulder. On the other hand, the positive trend of the answers to both the NASA-TLX and the single device Likert-Scale suggests that good acceptability and intuitiveness have already been achieved by both the ErgoTac and the CUFF. 

\begin{table*}[t]
\begin{center}
\caption{7-Points Likert-Scale. Queestions are divided in three groups: E-Q\# and C-Q\# for the questions provided after the ErgoTac and the CUFF block, respectively, F-Q\# for the questions provided in the final questionnaire.\label{Tab::LT}}
\rowcolors{2}{Gray}{white}
\begin{tabularx}{\textwidth}{rX|cc}
\toprule
\multicolumn{2}{c|}{Question}&Mean&STD\\
\midrule
E-Q1 & The ErgoTac device interfered with my movements. & 2.58 & 2.23\\
E-Q2 & Tactile feedback was intuitive when provided through the ErgoTac device on the upper limb. & 5.92 & 1.51\\
E-Q3 & It was difficult to interpret the feedback from the ErgoTac device on the lower limb. & 4.75 & 1.48\\
E-Q4 & It was easy to distinguish and interpret the inputs from the ErgoTac device on the two locations. & 4.58&1.78\\
E-Q5 & I felt tired while using the ErgoTac device. & 2.50 & 1.38\\
E-Q6 & The stimuli provided by the ErgoTac device were easily perceivable during the task. & 4.17 & 1.64\\
\midrule
C-Q1 & The CUFF device interfered with my movements. & 3.67 & 2.15\\
C-Q2 & Tactile feedback was intuitive when provided through the CUFF device on the upper limb. & 4.67 & 1.23\\
C-Q3 & It was difficult to interpret the feedback from the CUFF device on the lower limb. & 3.50 & 1.68\\
C-Q4 & It was easy to distinguish and interpret the inputs from the CUFF device on the two locations. & 4.50 & 1.17\\
C-Q5 & I felt tired while using the CUFF device. & 2.58 & 1.31\\
C-Q6 & The stimuli provided by the CUFF device were easily perceivable during the task. & 4.83 & 1.11\\
\midrule
F-Q1 & I had the feeling of performing better while receiving feedback by the ErgoTac compared to the CUFF on the upper limb. & 5.34 & 1.83\\
F-Q2 & I had the perception of performing faster while using the CUFF device compared to the ErgoTac device on the upper limb. & 3.58 & 2.24\\
F-Q3 & I had the feeling of performing better while receiving feedback by the CUFF compared to the ErgoTac on the lower limb. & 5.0 & 1.81\\
F-Q4 & I had the perception of performing faster while using the ErgoTac device compared to the CUFF device on the lower limb. & 3.25 & 1.71\\
F-Q5 & It was easier to follow the double feedback while using the CUFF device compared to the ErgoTac device. & 4.0 & 1.28\\
F-Q6 & I did not use any visual information to achieve the task. & 6.5 & 1.45\\
F-Q7 & The noise coming from the actuators helped me during the task. & 4.25 & 2.30\\
F-Q8 & I did not feel tired in the end of the experiment. & 4.50 & 1.83\\ \bottomrule
\end{tabularx}
\end{center}
\vspace{-0.5cm}
\end{table*}

In summary, for a task involving the upper limb, ErgoTac showed to be a better solution as a postural correction device. In addition, the hardware advantages of ErgoTac over the CUFF due to its reduced weight, limited size and lower power consumption cannot be deprecated. 
On the other hand, considering the promising results of the CUFF for the knee, it could be ideal for knee-focused applications. 

In whole-body applications where power consumption is not an issue, adopting a multi-modal solution with  ErgoTac on the shoulder and CUFF on the knee could be a reasonable possibility. 
In any case, these outcomes are also subject to the current state of the hardware. Hence, considering that both devices are prototypes so far and not industrial solutions yet, further studies are to be carried out considering future implementations and improvements of the devices. Finally, it is worth noticing that these results yield for the specific tasks under investigation. Future tests with different and more complex experimental tasks are needed to devise more robust conclusions.

\section{Conclusions and Future Works}
\label{Sec::Conclusions}
In this work, two previously proposed haptic devices, the CUFF and ErgoTac, were compared in their capability of providing single and multi-joint guidance to the users. 
Both devices show good acceptability and intuitiveness.
For the single-joint guidance of the shoulder, ErgoTac demonstrated significantly higher performance and it seems to be generally better perceived by the subjects. 
Due to its optimised wearability and extremely low power consumption, ErgoTac confirmed to be a good solution for the postural correction of the upper limb. 
On the other hand, the CUFF showed more positive results with the knee, also under the subjective point of view of the participants. 

This comparative study has been propaedeutic to the optimal integration of the two devices as an assistive apparatus to warn the users about inappropriate postures and guide them toward more ergonomic configurations. In this direction, future studies will focus on the evaluation of the two devices in guiding the users during industrial-like tasks, considering also a dual-task paradigm with distraction tasks. Based on the findings, for applications where energy consumption is not an issue, the combination of the ErgoTac guidance on the upper limb with the CUFF assistance on the lower limb will be examined. 
As previously mentioned, these results are based on a preliminary assessment of the systems. A more exhaustive characterization is needed to devise robust conclusions, which will include different experimental tasks also with longer duration. The usage of the CUFF device, in a multi-modal configuration with the Ergotac, could  minimize the amount of vibrotactile stimuli delivered to the users over long periods, hence preventing possible adaptation/saturation of the receptive channels \cite{fani2021multi}. The cognitive load of this configuration will be also investigated.



\bibliographystyle{IEEEtran}
\bibliography{bibliogra}

\begin{thebibliography}{10}
\providecommand{\url}[1]{#1}
\csname url@samestyle\endcsname
\providecommand{\newblock}{\relax}
\providecommand{\bibinfo}[2]{#2}
\providecommand{\BIBentrySTDinterwordspacing}{\spaceskip=0pt\relax}
\providecommand{\BIBentryALTinterwordstretchfactor}{4}
\providecommand{\BIBentryALTinterwordspacing}{\spaceskip=\fontdimen2\font plus
\BIBentryALTinterwordstretchfactor\fontdimen3\font minus
  \fontdimen4\font\relax}
\providecommand{\BIBforeignlanguage}[2]{{%
\expandafter\ifx\csname l@#1\endcsname\relax
\typeout{** WARNING: IEEEtran.bst: No hyphenation pattern has been}%
\typeout{** loaded for the language `#1'. Using the pattern for}%
\typeout{** the default language instead.}%
\else
\language=\csname l@#1\endcsname
\fi
#2}}
\providecommand{\BIBdecl}{\relax}
\BIBdecl

\bibitem{EUOSHA2019}
J.~de~Kok, P.~Vroonhof, J.~Snijders, G.~Roullis, M.~Clarke, K.~Peereboom,
  P.~van Dorst, and I.~Isusi, ``Work-related musculoskeletal disorders:
  prevalence, costs and demographics in the eu,'' EU-OSHA European Agency for
  Safety and Health at Work, Tech. Rep., 2019.

\bibitem{james2018global}
S.~James, D.~Abate, K.~Abate, S.~Abay, C.~Abbafati, N.~Abbasi, H.~Abbastabar,
  F.~Abd-Allah, J.~Abdela, A.~Abdelalim \emph{et~al.}, ``Global, regional, and
  national incidence, prevalence, and years lived with disability for 354
  diseases and injuries for 195 countries and territories, 1990--2017: a
  systematic analysis for the global burden of disease study 2017,'' \emph{The
  Lancet}, vol. 392, no. 10159, pp. 1789--1858, 2018.

\bibitem{da2010risk}
B.~da~Costa and E.~Vieira, ``Risk factors for work-related musculoskeletal
  disorders: a systematic review of recent longitudinal studies,''
  \emph{American journal of industrial medicine}, vol.~53, no.~3, pp. 285--323,
  2010.

\bibitem{punnett2004work}
L.~Punnett and D.~H. Wegman, ``Work-related musculoskeletal disorders: the
  epidemiologic evidence and the debate,'' \emph{Journal of electromyography
  and kinesiology}, vol.~14, no.~1, pp. 13--23, 2004.

\bibitem{bossomaier2010scientific}
T.~Bossomaier, A.~G. Bruzzone, A.~Cimino, F.~Longo, and G.~Mirabelli,
  ``Scientific approaches for the industrial workstations ergonomic design: A
  review,'' in \emph{European Conference on Modelling and Simulation
  (ECMS)}.\hskip 1em plus 0.5em minus 0.4em\relax ECMS, 2010, pp. 1--11.

\bibitem{ben2002ergonomic}
I.~Ben-Gal and J.~Bukchin, ``The ergonomic design of workstations using virtual
  manufacturing and response surface methodology,'' \emph{Iie Transactions},
  vol.~34, no.~4, pp. 375--391, 2002.

\bibitem{lim2015development}
D.~Lim, W.~Kim, H.~Lee, H.~Kim, K.~Shin, T.~Park, J.~Lee, and C.~Han,
  ``Development of a lower extremity exoskeleton robot with a
  quasi-anthropomorphic design approach for load carriage,'' in \emph{2015
  IEEE/RSJ International Conference on Intelligent Robots and Systems
  (IROS)}.\hskip 1em plus 0.5em minus 0.4em\relax IEEE, 2015, pp. 5345--5350.

\bibitem{peternel2014teaching}
L.~Peternel, T.~Petri{\v{c}}, E.~Oztop, and J.~Babi{\v{c}}, ``Teaching robots
  to cooperate with humans in dynamic manipulation tasks based on multi-modal
  human-in-the-loop approach,'' \emph{Autonomous robots}, vol.~36, no.~1, pp.
  123--136, 2014.

\bibitem{evrard2009teaching}
P.~Evrard, E.~Gribovskaya, S.~Calinon, A.~Billard, and A.~Kheddar, ``Teaching
  physical collaborative tasks: object-lifting case study with a humanoid,'' in
  \emph{2009 9th IEEE-RAS International Conference on Humanoid Robots}.\hskip
  1em plus 0.5em minus 0.4em\relax IEEE, 2009, pp. 399--404.

\bibitem{diefenbach2021improving}
H.~Diefenbach, N.~Erlemann, A.~Lunin, E.~H. Grosse, K.-O. Schocke, and C.~H.
  Glock, ``Improving processes and ergonomics at air freight handling agents: a
  case study,'' \emph{International Journal of Logistics Research and
  Applications}, pp. 1--22, 2021.

\bibitem{goomas2010ergonomics}
D.~T. Goomas and P.~H. Yeow, ``Ergonomics improvement in a harsh environment
  using an audio feedback system,'' \emph{International Journal of Industrial
  Ergonomics}, vol.~40, no.~6, pp. 767--774, 2010.

\bibitem{pacchierotti2017wearable}
C.~Pacchierotti, S.~Sinclair, M.~Solazzi, A.~Frisoli, V.~Hayward, and
  D.~Prattichizzo, ``Wearable haptic systems for the fingertip and the hand:
  taxonomy, review, and perspectives,'' \emph{IEEE transactions on haptics},
  vol.~10, no.~4, pp. 580--600, 2017.

\bibitem{barontini}
F.~Barontini, M.~G. Catalano, L.~Pallottino, B.~Leporini, and M.~Bianchi,
  ``Integrating wearable haptics and obstacle avoidance for the visually
  impaired in indoor navigation: A user-centered approach,'' \emph{IEEE Trans.
  on Haptics}, vol.~14, no.~1, pp. 109--122, 2021.

\bibitem{baldi2020hand}
T.~L. Baldi, N.~d’Aurizio, and D.~Prattichizzo, ``Hand guidance using
  grasping metaphor and wearable haptics,'' in \emph{2020 IEEE Haptics
  Symposium (HAPTICS)}.\hskip 1em plus 0.5em minus 0.4em\relax IEEE, 2020, pp.
  961--967.

\bibitem{pezent2018separating}
E.~Pezent, S.~Fani, J.~Bradley, M.~Bianchi, and M.~K. O'Malley, ``Separating
  haptic guidance from task dynamics: A practical solution via cutaneous
  devices,'' in \emph{2018 IEEE Haptics Symposium (HAPTICS)}.\hskip 1em plus
  0.5em minus 0.4em\relax IEEE, 2018, pp. 20--25.

\bibitem{rotella2012hapi}
M.~F. Rotella, K.~Guerin, X.~He, and A.~M. Okamura, ``Hapi bands: a haptic
  augmented posture interface,'' in \emph{2012 IEEE Haptics Symposium
  (HAPTICS)}.\hskip 1em plus 0.5em minus 0.4em\relax IEEE, 2012, pp. 163--170.

\bibitem{kim2021directional}
W.~Kim, V.~R. Garate, J.~M. Gandarias, M.~Lorenzini, and A.~Ajoudani, ``A
  directional vibrotactile feedback interface for ergonomic postural
  adjustment,'' \emph{IEEE Transactions on Haptics}, 2021.

\bibitem{fani2021multi}
S.~Fani, S.~Ciotti, and M.~Bianchi, ``Multi-cue haptic guidance through
  wearables for enhancing human ergonomics,'' \emph{IEEE Transactions on
  Haptics}, 2021.

\bibitem{kim2018ergotac}
W.~Kim, M.~Lorenzini, K.~Kap{\i}c{\i}o{\u{g}}lu, and A.~Ajoudani, ``Ergotac: A
  tactile feedback interface for improving human ergonomics in workplaces,''
  \emph{IEEE Robotics and Automation Letters}, vol.~3, no.~4, pp. 4179--4186,
  2018.

\bibitem{fani2018simplifying}
S.~Fani, S.~Ciotti, M.~G. Catalano, G.~Grioli, A.~Tognetti, G.~Valenza,
  A.~Ajoudani, and M.~Bianchi, ``Simplifying telerobotics: wearability and
  teleimpedance improves human-robot interactions in teleoperation,''
  \emph{IEEE Robotics \& Automation Magazine}, vol.~25, no.~1, pp. 77--88,
  2018.

\bibitem{casini2015design}
S.~Casini, M.~Morvidoni, M.~Bianchi, M.~Catalano, G.~Grioli, and A.~Bicchi,
  ``Design and realization of the cuff-clenching upper-limb force feedback
  wearable device for distributed mechano-tactile stimulation of normal and
  tangential skin forces,'' in \emph{2015 IEEE/RSJ International Conference on
  Intelligent Robots and Systems (IROS)}.\hskip 1em plus 0.5em minus
  0.4em\relax IEEE, 2015, pp. 1186--1193.

\bibitem{pezent2019spatially}
E.~Pezent, S.~Fani, J.~Clark, M.~Bianchi, and M.~K. O'Malley, ``Spatially
  separating haptic guidance from task dynamics through wearable devices,''
  \emph{IEEE transactions on haptics}, vol.~12, no.~4, pp. 581--593, 2019.

\bibitem{hart1988development}
S.~G. Hart and L.~E. Staveland, ``Development of nasa-tlx (task load index):
  Results of empirical and theoretical research,'' in \emph{Advances in
  psychology}.\hskip 1em plus 0.5em minus 0.4em\relax Elsevier, 1988, vol.~52,
  pp. 139--183.

\end{thebibliography}

\end{document}